\documentclass[letterpaper, 10 pt, conference]{ieeeconf}  % Comment this line out if you need a4paper

\IEEEoverridecommandlockouts                              % This command is only needed if 
\usepackage{graphics} % for pdf, bitmapped graphics files
\usepackage{epsfig} % for postscript graphics files
\usepackage{times} % assumes new font selection scheme installed
\usepackage{amsmath} % assumes amsmath package installed
\usepackage{amssymb}  % assumes amsmath package installed

\usepackage[ruled,vlined,linesnumbered]{algorithm2e}
\usepackage{xargs}    % for custom commands
\usepackage{textcomp}
\usepackage{multirow}
\usepackage{lipsum}
\usepackage{subcaption}
\usepackage{xcolor}
\usepackage{bm}
\usepackage{cite}
\usepackage{float}
\usepackage{graphicx}
\usepackage[]{todonotes}    % todo notes. add [disable] to disable
\usepackage{framed, color} % for colored frames around text
\usepackage[normalem]{ulem}
\usepackage{algorithm2e}

\title{Meta Reinforcement Learning for Sim-to-real Domain Adaptation}

\author{Karol Arndt$^{1}$, Murtaza Hazara$^{1}$, Ali Ghadirzadeh$^{1,2}$, Ville Kyrki$^{1}$
\thanks{*This work was financially supported by Academy of Finland grant 313966 and Business Finland grant 3338/31/2017. We also gratefully acknowledge the support of NVIDIA Corporation with the donation of the Titan Xp GPU used for this research.}% <-this % stops a space
\thanks{$^{1}$Aalto University, Espoo, Finland
        {\tt\small first.last@aalto.fi}}%
\thanks{$^{2}$KTH Royal Institute of Technology, Stockholm, Sweden}
}

\definecolor{karolsfavouritecolour}{HTML}{7100BC}
\newcommandx{\karol}[2][1=]{\todo[backgroundcolor=karolsfavouritecolour!50,inline,#1]{K: #2}\noindent}
\newcommandx{\karinl}[2][1=]{\textcolor{karolsfavouritecolour}{\textbf{#2}}\noindent}
\newcommand\karweg{\bgroup\markoverwith{\textcolor{karolsfavouritecolour}{\rule[0.5ex]{2pt}{2pt}}}\ULon}

\SetKwFor{RepTimes}{repeat}{times}{end}

\begin{document}
\maketitle

% Actual content

\begin{abstract}
Modern reinforcement learning methods suffer from low sample efficiency and unsafe exploration, making it infeasible to train robotic policies entirely on real hardware.
In this work, we propose to address the problem of sim-to-real domain transfer by using meta learning to train a policy that can adapt to a variety of dynamic conditions, and using a task-specific trajectory generation model to provide an action space that facilitates quick exploration.
We evaluate the method by performing domain adaptation in simulation and analyzing the structure of the latent space during adaptation.
We then deploy this policy on a KUKA LBR 4+ robot and evaluate its performance on a task of hitting a hockey puck to a target.
Our method shows more consistent and stable domain adaptation than the baseline, resulting in better overall performance.

\end{abstract}

% Introduction
\section{Introduction}
\label{sec:intro}

In recent years, we have witnessed a tremendous progress in reinforcement learning research, accompanied by its growing application in robotics.
%However, despite many promising results, training policies for robots still often requires months, if not years, of experience, 
%making the training process infeasible to perform entirely on physical hardware~\cite{levine2017learning}.
Reinforcement learning, however, requires vast amounts of training data, which can be relatively costly to provide in robotics~\cite{levine2017learning}, in contrast to applications like computer games~\cite{mnih2013playing,mnih2015human}.
Apart from reliance on large amounts of data, with most methods the training process involves random exploratory actions, which can be unpredictable and potentially unsafe both to the operational environment and to the robot itself.

A promising solution to these problems lies in using physics simulators, such as MuJoCo~\cite{Todorov2012MuJoCoAP} or Flex~\cite{liang18flex}, to reduce training time and mitigate the risk of hardware damage~\cite{chebotar2018closing,tan2018sim,hamalainen2019affordance}.
However, directly deploying the trained model on physical hardware still requires an accurate match between the simulation and real-world, which may be impossible to achieve even after tedious tuning of simulation parameters, because the simulation may not model some of the physical phenomena present in real world. 
%To address these issues, methods based on \textit{domain adaptation} have been proposed~\cite{peng2018sim,transferMurtaza}.
%However, recent developments in the field of meta learning have made it a compelling, yet still understudied,  approach to this problem.
As an alternative to carefully tuning the simulator, a model trained on imprecise dynamics can be adapted to the real world environment to make up for potential modelling inaccuracies~\cite{peng2018sim,transferMurtaza}.
Recent developments in the field of meta learning made it a compelling, yet still understudied, approach to this problem.

\begin{figure}
    \centering
    \begin{subfigure}{0.49\linewidth}
    \centering
    \includegraphics[width=\linewidth]{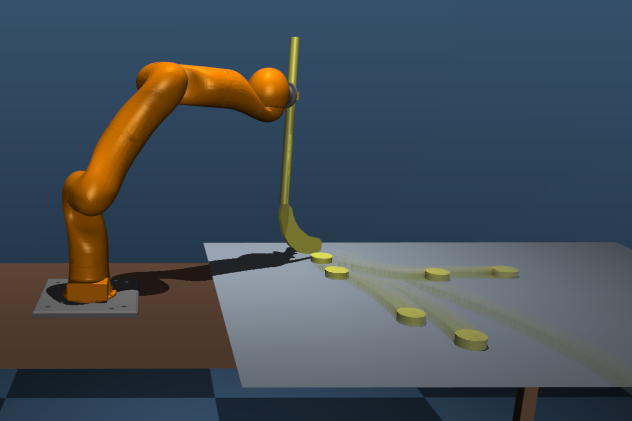}
    \caption{}
    \label{fig:blur_sim}
    \end{subfigure}
    \begin{subfigure}{0.49\linewidth}
    \centering
    \includegraphics[width=\linewidth]{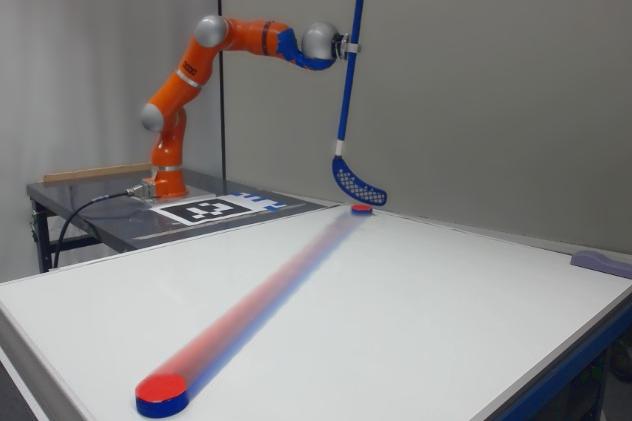}
    \caption{}
    \label{fig:blur_real}
    \end{subfigure}
    \caption{Randomized dynamic properties lead to large changes in the behaviour of the system. We train a model to adapt to large variations in simulation~(\subref{fig:blur_sim}), and deploy it on a physical robot~(\subref{fig:blur_real}).}
    \label{fig:front_figure}
\end{figure}

In this work, we propose a novel method for domain adaptation using meta learning, or \textit{learning to learn}.
As opposed to most common machine learning problem formulations, where the goal is to train a model to excel in one particular task, the basic principle in meta learning is to train models which are good at adapting to new tasks or situations.

We consider tasks that are heavily dependent on the dynamic parameters of the environment.
Such tasks cannot be transferred from simulation to reality in a zero-shot manner and require data from the physical system to be used for domain adaptation or system identification.
We combine gradient-based meta learning with generative models for trajectories to explicitly train a policy to adapt to a wide range of randomized dynamics in simulation, illustrated in Figure~\ref{fig:blur_sim}.
The trained model is then deployed on a physical setup, shown in Figure~\ref{fig:blur_real}, quickly adapting to new conditions.
The method's feasibility for sim-to-real transfer is demonstrated on a task where the goal is to shoot a hockey puck to a target location under unknown friction.
%The impact of dynamic parameters on the behaviour of a physical system is illustrated in Figure~\ref{fig:blur_sim} --- the same trajectory executed by the robot will result in the puck stopping at a wide variety of different locations, depending on the friction between the puck and the surface.
%We demonstrate the feasibility of our method using a similar hockey puck experimental setup, shown in Figure~\ref{fig:blur_real}, where the goal is to shoot a hockey puck to a target location under unknown friction.
We show that, after a small number of trials, our system is able to adapt both to new conditions in simulation and to the dynamic parameters of the physical system, improving the performance in cases where simple domain adaptation methods fail, or result in unstable policy updates.

The contributions of our work are (1) demonstrating that gradient-based meta learning results in predictable and consistent domain adaptation, and thus is a suitable approach for simulation to reality (sim-to-real) transfer of robotic policies under uncertain dynamics, and (2) combining a meta learned policy with latent variable generative models to represent motor trajectories leads to a safe and low-dimensional exploration space.

% Related work
\section{Related work}
\label{sec:related}
In this section, we provide an overview of previous work related to meta learning and sim-to-real transfer.

\subsection{Meta learning}
The general idea of meta learning was first described by Schmidhuber~\cite{schmidhuber:1987:srl}, whose early work on the topic pioneered the use of meta learning with neural network models~\cite{Schmidhuber93neural}, as well as its application in reinforcement learning~\cite{Schmidhuber1998reinforcement}.

More recently, two principal families of meta learning algorithms have been introduced.
% Meta learning can be used for task adaptation in two principal ways: 
First, memory can be embedded as a part of the learned structure, causing the network to adapt as data gets passed through it.
Such behaviour can be accomplished by recurrent architectures~\cite{Ravi2017OptimizationAA, Duan2017RL2FR} or by using an additional set of plastic weights~\cite{Miconi2018Differentiable}.

Second, parameters of the network can be optimized such that they provide a good starting point for further adaptation.
This is the case in model-agnostic meta learning (MAML)~\cite{finn17maml} which explicitly optimizes the model performance after a number of adaptation steps. 
MAML has been demonstrated to achieve good performance in tasks such as few-shot classification and reinforcement learning, including more complex tasks, such as robot simulations~\cite{brockman16gym}.
Multiple improvements to the method were later introduced by various authors~\cite{Antoniou2018HowTT, stadie18emaml}.

\subsection{Sim-to-real domain transfer}

Zero shot transfer refers to learning a policy that does not need to be adapted in the target domain. 
A common approach for zero-shot transfer is domain randomization, which exposes the model to a variety of conditions, so as to make the model robust to modelling inaccuracies in these aspects.
The idea can be applied both to perception~\cite{tobin2017domain,hamalainen2019affordance,Sadeghi16real,sadeghi17sim2real} and to the dynamics of the system~\cite{OpenAI2018LearningDI,petrik2019feedback}. Domain randomization may, however, be insufficient since a single policy that performs well across the domain might not exist. 

One solution is to build a more accurate simulation model for the particular environment, either by thorough measurements~\cite{tan2018sim} or by interweaving simulation rollouts with real robot samples and optimizing the simulation~\cite{chebotar2018closing} or the policy~\cite{wulfmeier17mutual}, such that the discrepancies are minimized.
This can be, however, costly and time-consuming, and requires access to physical hardware at the time of training.
As a solution, memory can be embedded as part of the network to encode previous states and actions, allowing the network to identify and respond to a variety of dynamic conditions~\cite{peng2018sim}.
%We take a similar approach for model-free policy adaptation; however, we focus on gradient-based meta learning methods.

The problem can also be approached from a different direction---the policy trained in simulation can also be directly used as a starting point for further adaptations in real world~\cite{transferMurtaza}. 
The initial parameters may, however, not be a good point for further adaptation.
Gradient-based meta learning methods are an appealing solution to this problem.
A method for stabilizing model-based reinforcement learning using gradient-based meta learning was proposed by Clavera~et~al.~\cite{clavera2018model} to address minor uncertainties in dynamics originating from lack of related training data. 
In contrast, in this work, we directly adapt the policy to a wide range of dynamic conditions using model-free methods, and additionally evaluate its performance on a physical system.

% Method
\section{Method}
\label{sec:method}
In this section, our method for sim-to-real transfer learning is introduced. First, we describe the necessary preliminaries related to domain adaptation using meta reinforcement learning, followed by the formal problem formulation.
We then describe the details of our approach for trajectory generation, domain adaptation, and meta-policy training.

\subsection{Preliminaries and problem statement}
A standard sequential decision making setup consists of an \textit{agent} interacting with an \textit{environment} in discrete timesteps.
At each timestep $t$, the agent takes an action $a_t$, causing the environment to change its state from $s_t$ to $s_{t+1}$.
Each state transition is accompanied by a corresponding reward $r(s_t, a_t)$ to assess the quality of the action.
This setup is a Markov decision process (MDP) with a set of states $s \in \mathcal{S}$, actions $a \in \mathcal{A}$, and transition probabilities between these states in response to each action~$p(s_{t+1} | s_{t}, a_{t})$.
The agent's actions are chosen according to a policy $\pi(a_t | s_t)$, which describes the probability of taking action~$a_t$ in state~$s_t$.
The objective of reinforcement learning then is to find the optimal policy, defined as the policy that maximizes the expected cumulative sum of rewards for a specific MDP.

In contrast to this formulation, meta reinforcement learning considers a set of MDPs, $\mathcal{M}$.
The goal is to find a learning algorithm that is able to efficiently learn optimal policies for all MDPs in $\mathcal{M}$ --- that is, to \textit{learn to learn} policies in $\mathcal{M}$.
In the domain adaptation scenario, we consider $\mathcal{M}$ to consist of MDPs sharing the same reward function $r(s, a)$, as well as the action and state spaces ($\mathcal{A}$ and $\mathcal{S}$), but varying in terms of state transition probabilities~$p(s_{t+1} | s_t, a_t)$.
We further assume that, for each MDP $M_k \in \mathcal{M}$, these transitions can be described by a set of dynamic parameters, further referred to as \textit{task} $\tau \in \boldsymbol{\tau}$.

In the context of meta learning, the domain adaptation problem can therefore be stated as follows:
for a set of Markov decision processes $\mathcal{M}$, described by tasks $\tau_i \in \boldsymbol{\tau}$, find a learning algorithm which, after performing $N$ adaptation steps under a new dynamic condition $\tau$, results in the optimal policy for these conditions, $\pi^*_{\theta,\tau}$.

\subsection{Trajectory generation}
\label{sec:trajgen}
In order to provide the policy with a low-dimensional, smooth action space which facilitates exploration, we train a generative model over a distribution of task-specific trajectories $u_{0:T} = g_{\phi}(z)$, where $u$ is a trajectory and $g_\phi$ the generative model parametrized by a latent variable $z$. This is similar to~\cite{ghadirzadeh2017deep} and \cite{hamalainen2019affordance}. 
%This approach is similar to, and draws inspiration from, the way how generative models for images are trained~\cite{kingma13vae, Higgins2017betaVAE}.

We obtain the generative model by training a variational autoencoder (VAE) on a set of trajectories which are suitable for the given task and safe to be executed on the physical robot.
A VAE consists of two parts---the encoder and the decoder.
The encoder outputs a probability distribution representing the low-dimensional latent representation of the input.
During training, a sample is drawn from this distribution and passed to the decoder, which reconstructs the original input based on the low-dimensional representation.
%This architecture causes the network to learn low-dimensional representation of the data, while promoting representations which are smooth and disentangled~\cite{Higgins2017betaVAE, burgess2018understanding, mathieu2019disentangling}.
The decoder part of the VAE, on its own, can be used to map vectors in the latent space to the output domain, which in our case represents useful trajectories.

This formulation allows us to train a policy for latent actions $z$, $\pi(z | s)$, effectively reducing the dimensionality of the action space, alleviating the problem of time complexity and allowing the model to focus on terminal rewards.
Effectively, this formulation results in faster training and safer on-policy domain adaptation.

\subsection{Domain adaptation}
\label{sec:domainadaptation}
\begin{figure}
    \centering
    \includegraphics[width=\linewidth]{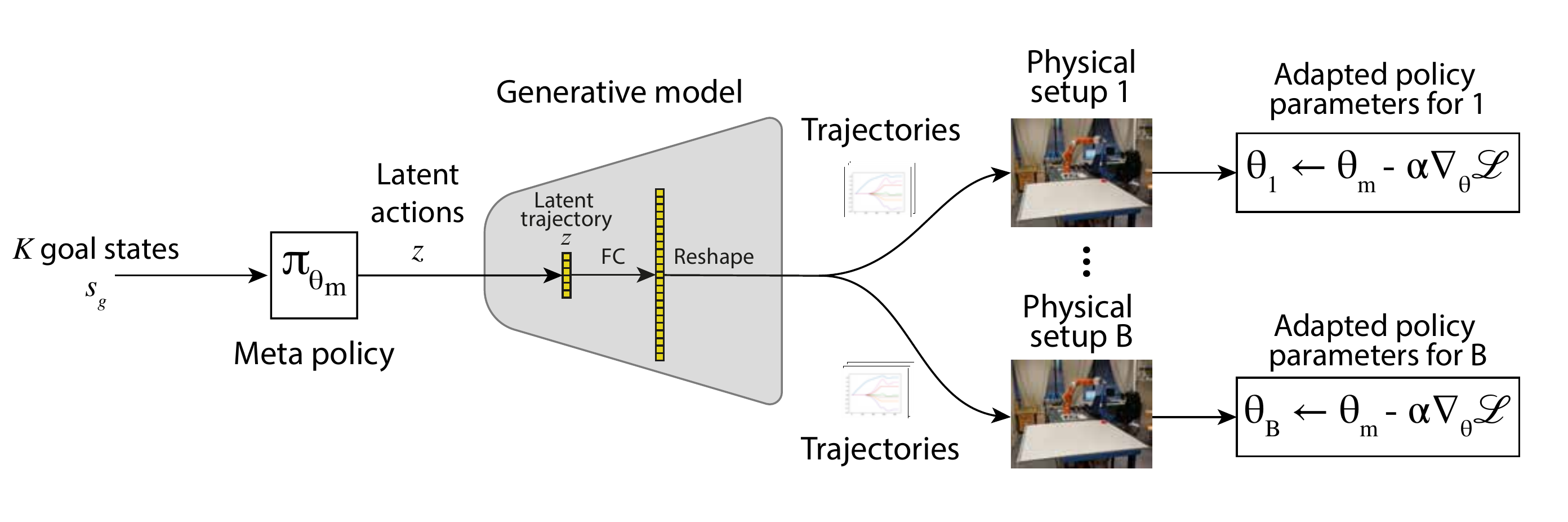}
    \caption{Overview of one step of the adaptation procedure.}
    \label{fig:adapt_overview}
    \vspace{-4pt}
\end{figure}

The goal of the domain adaptation step is to adjust the policy parameters in such a way that the policy's performance will improve for the current dynamic conditions.
This process is outlined in Figure~\ref{fig:adapt_overview} and in Algorithm~\ref{alg:adaptation}.
For clarity, Figure~\ref{fig:adapt_overview} shows only a single adaptation step ($N=1$).

\SetKwInput{KwData}{Input}
\begin{algorithm}
\SetAlgoLined
\KwData{trained generator $g_\phi$ and meta policy $\pi_{\theta_0}$}
\KwResult{Adapted policy parameters $\theta_N$}
 \RepTimes{$n:=N$}{
     \RepTimes{$k:=K$}{
         get goal state $s_g$\; \label{alg1:get_goal}
         sample $z_k \sim \pi_{\theta_n}(z | s_g)$\;  \label{alg1:sample_z}
         generate trajectory $u_{0:T} := g_\phi(z_k)$\;  \label{alg1:gen_traj}
         execute $u_{0:T}$, save $s_g$, $z_k$ and the reward $r_k$\;  \label{alg1:exec_traj}
     }
     normalize rewards $\bar{\mathbf{r}} = \frac{\mathbf{r} - mean(\mathbf{r})}{std dev(\mathbf{r})}$\;  \label{alg1:norm}
     calculate loss $\mathcal{L} := -\frac{1}{K} \sum_k^K \tilde{r}_k \log \pi(z_k | s_k)$\;  \label{alg1:loss}
     update policy parameters $\theta_{n} := \theta_{n-1} - \alpha \nabla_{\theta_{n-1}} \mathcal{L}$\;  \label{alg1:update}
 }
 \caption{Policy adaptation}
 \label{alg:adaptation}
\end{algorithm}

The adaptation begins by sampling a random goal state from the environment and passing it to the current policy (step~\ref{alg1:get_goal}).
The policy returns a latent action distribution $\pi(z | s)$, from which a latent vector $z$ is sampled and passed to the generative model $g_\phi$ to construct the corresponding trajectory (steps~\ref{alg1:sample_z} and~\ref{alg1:gen_traj}).
The constructed trajectory is then executed by the robot and the state, action and reward are stored (step~\ref{alg1:exec_traj}).
This process is repeated $K$ times.

After $K$ rollouts from the policy are collected, the policy is adapted by updating its parameters using vanilla policy gradient (steps~\ref{alg1:norm} to~\ref{alg1:update}).
The whole process is repeated $N$ times.
Building on this, we will now describe the meta policy training procedure that provides the input meta policy for the policy adaptation.

\subsection{Training the meta-policy}
\begin{figure}
    \centering
    \includegraphics[width=\linewidth]{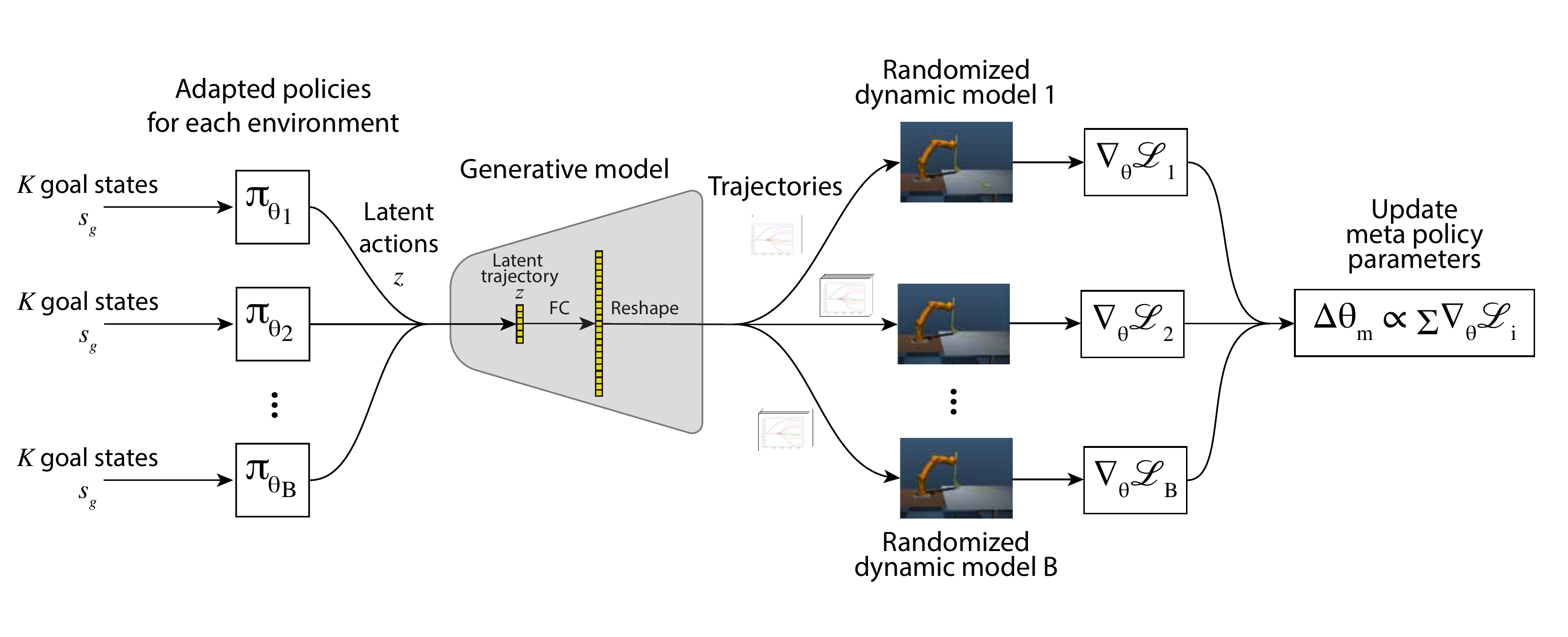}
    \caption{Overview of the meta training procedure, starting from the adapted policies for each environment.}
    \label{fig:meta_overview}
\end{figure}

The objective of training the meta policy is to find the optimal meta parameters $\theta_m$, which result in fast adaptation to new dynamic conditions. 
We propose a process similar to MAML which is illustrated in Figure~\ref{fig:meta_overview} and outlined in Algorithm~\ref{alg:metatraining}.
The process starts with sampling a batch of tasks $\tau_i \sim p(\tau)$ (step~\ref{alg2:sample_batch}).
Each task represents a new environment with different, randomized dynamics.
For each of the environments, the agent starts with the meta policy and performs $N$ adaptation steps, each using $K$ rollouts from the policy, as described in Section~\ref{sec:domainadaptation} (step~\ref{alg2:adapt}).
After the last adaptation step, the agent collects $K$ rollouts using the final adapted policy (step~\ref{alg2:get_data}).
This data is, in turn, used to update the final adapted policy  in step~\ref{alg2:update}. However, instead of directly updating the parameters of the final adapted policy, the gradients are backpropagated through all $N$ update steps, all the way back to the parameters of the original meta policy.
This update can be performed using any model free reinforcement learning algorithm.

This process results in a model that is explicitly optimized to maximize the expected cumulative return after $N$ adaptation steps, in contrast to training a single universal policy to perform well on all tasks at the same time.

\begin{algorithm}
\SetAlgoLined
\KwData{Trained generator $g_\phi$}
\KwResult{Meta policy parameters $\theta_m$}
 randomly initialize meta parameters $\theta_m$\;
 \While{not converged}{
     \RepTimes{$b:=B$}{ \label{alg2:sample_batch}
     Sample task $\tau_b$\; \label{alg2:sample_task}
    perform $N$ adaptation steps as in Algorithm~\ref{alg:adaptation} \; \label{alg2:adapt}
    collect $K$ samples using $\pi_{\theta_N,b}$ \; \label{alg2:get_data}
 }
 Update $\theta_m$ to improve performance of $\pi_{\theta_N,b}$ for all $\tau_b$\; \label{alg2:update}
 }
 \caption{Meta policy training}
 \label{alg:metatraining}
\end{algorithm}
 \vspace{-5pt}

% Experimental evaluation
\section{Experimental Evaluation}
\label{sec:result}
In this section, we describe the experimental setup and the architecture of our models, together with their training procedures and the description of a baseline method. Then, we present the adaptation results obtained in simulation and on a physical setup, comparing the performance of our method to the presented baseline.

\subsection{Experimental setup}
\label{sec:setup}
\begin{figure}
\vspace{6pt}
    \centering
    \begin{subfigure}{0.48\linewidth}
    \centering
    \includegraphics[width=\linewidth]{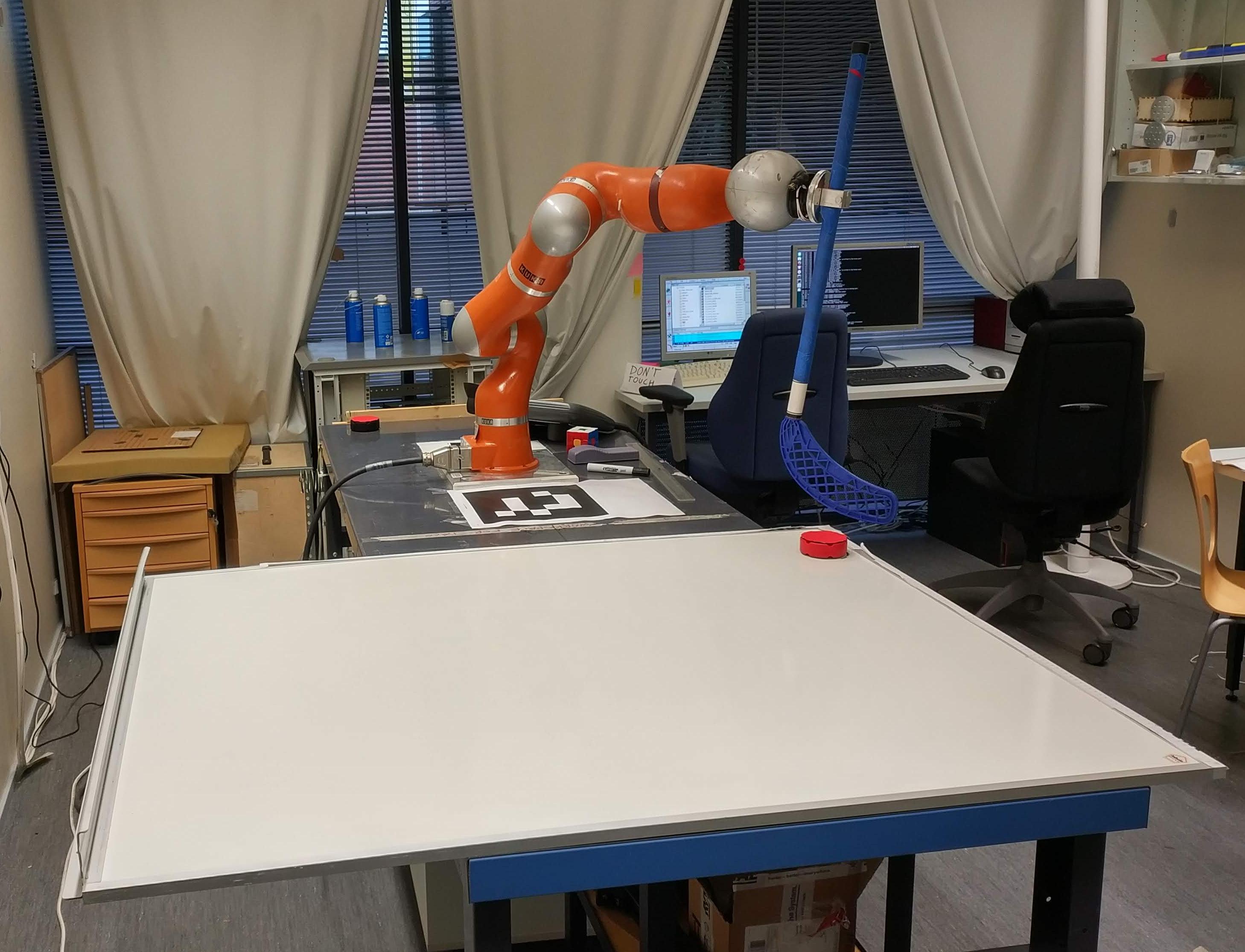}
    \caption{}
    \label{fig:hockey_setu}
    \end{subfigure}
    \begin{subfigure}{0.48\linewidth}
    \centering
    \includegraphics[width=\linewidth]{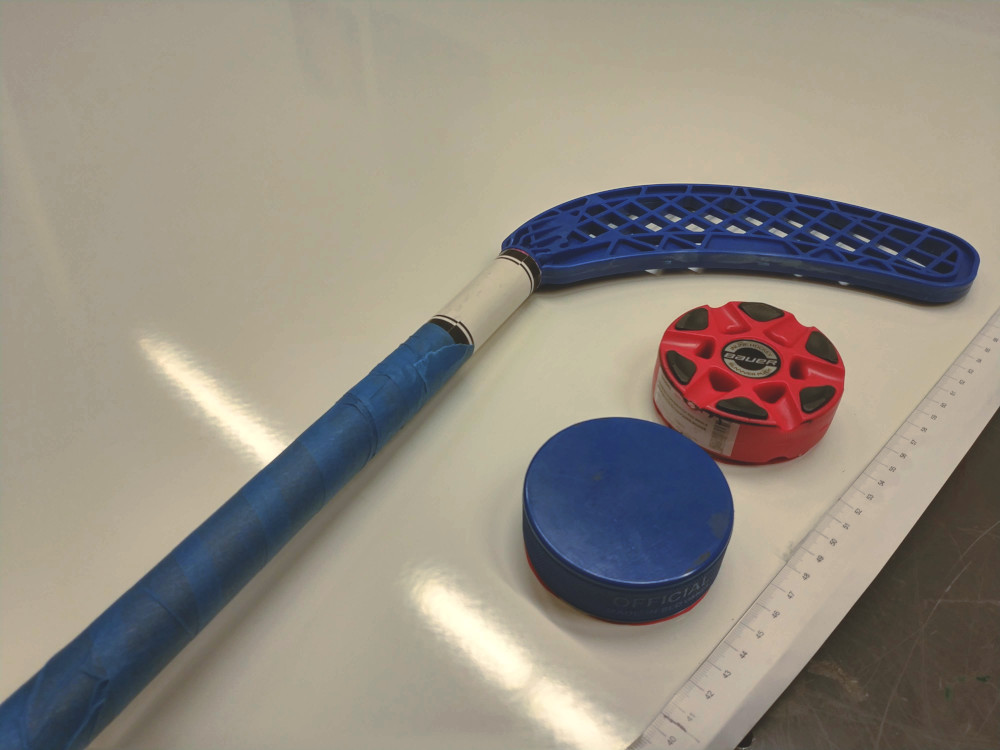}
    \caption{}
    \label{fig:hockey_pucks}
    \end{subfigure}
    \label{fig:phys}
    \caption{The hockey puck experimental setup~(\subref{fig:hockey_setu}) and the tools used for the experiments~(\subref{fig:hockey_pucks})}
\end{figure}

The hockey puck setup consists of a KUKA LBR 4+ robot equipped with a floorball stick, as illustrated in Figure~\ref{fig:hockey_setu}.
The robot uses the stick to hit a hockey puck on a flat, low friction surface, such that the puck lands in a target location.
The friction between the surface and the puck has crucial impact on the movements of the puck, forcing the policy to learn how to operate in different friction conditions.
We use two different hockey pucks, as shown in Figure~\ref{fig:hockey_pucks}: an ice hockey puck (low friction; blue), and an inline hockey puck (higher friction; red).
Since all contact points of the inline hockey puck are located close to the edge, it also has noticeably higher torsional friction than the ice hockey puck.
The position of the pucks is measured by a camera mounted on the ceiling above the whiteboard.
The target range of size 50cm x 30cm is located close to the center of the whiteboard.

We constructed a corresponding simulation setup in MuJoCo~\cite{Todorov2012MuJoCoAP}.
During training in simulation, we randomize the mass of the puck, the five friction parameters between the puck and the surface, and the starting position to account for possible misalignments between the real setup and the simulation.
The randomization parameters are presented in Table~\ref{tab:randomizations}.
We use uniform distributions for the dynamic parameters and a normal distribution for the starting position noise.

\begin{table}[]
    \centering
    \caption{Randomized parameters}
    \begin{tabular}{c|c|c}
        Parameter & Minimum & Maximum \\
        \hline
        $x$ linear friction ($\mu_x$) & 0.15 & 0.95 \\
        $y$ linear friction ($\mu_y$) & 0.7$\mu_x$ & 1.3$\mu_x$ \\
        Torsional friction ($\mu_\tau$) & 0.001 & 0.05 \\
        Rotational friction $x$ ($\mu_{rx}$) & 0.01 & 0.3 \\
        Rotational friction $y$ ($\mu_{ry}$) & 0.01 & 0.3 \\
        Puck mass & 50g & 500g  \\
        Initial puck position & \multicolumn{2}{c}{$\epsilon \sim \mathcal{N}(0, 0.02)$}
    \end{tabular}
    \label{tab:randomizations}
\end{table}

Each parameter of the system is randomized separately, except for $\mu_y$, which is randomized in relatively to $\mu_x$.
The logic behind the use of anisotropic friction was based on an observation that under the same robot trajectories, puck movement directions on the physical setup are noticeably different from the simulation.
We presume that this is caused by contact modeling inaccuracies and unmodeled effects such as the elasticity of the hockey blade and unevenness of the surface.
Instead of fine-tuning the simulated behaviour to be closer to reality, we aimed to make up for these inaccuracies with additional randomizations.

\subsection{Training the generative trajectory model}
To train the hockey puck trajectory generation model, we generated 7371 trajectories consisting of 17 waypoints of a cubic spline.
These trajectories were obtained by moving the robot from the starting position to the proximity of the puck, making a swing, and moving the hockey blade past the puck.
The strength of the swing and the orientation of the blade were randomized in order to generate a variety of trajectories with different hitting strengths and hitting angles.

These trajectory waypoints were then used to train the trajectory generation model.
We used a 2 dimensional latent space throughout the experiments.
Similarly to~\cite{hamalainen2019affordance}, we increased the value of $\beta$ during training from $10^{-7}$ to $10^{-3}$.

\begin{figure}
\vspace{6pt}
    \centering
    \includegraphics[width=\linewidth]{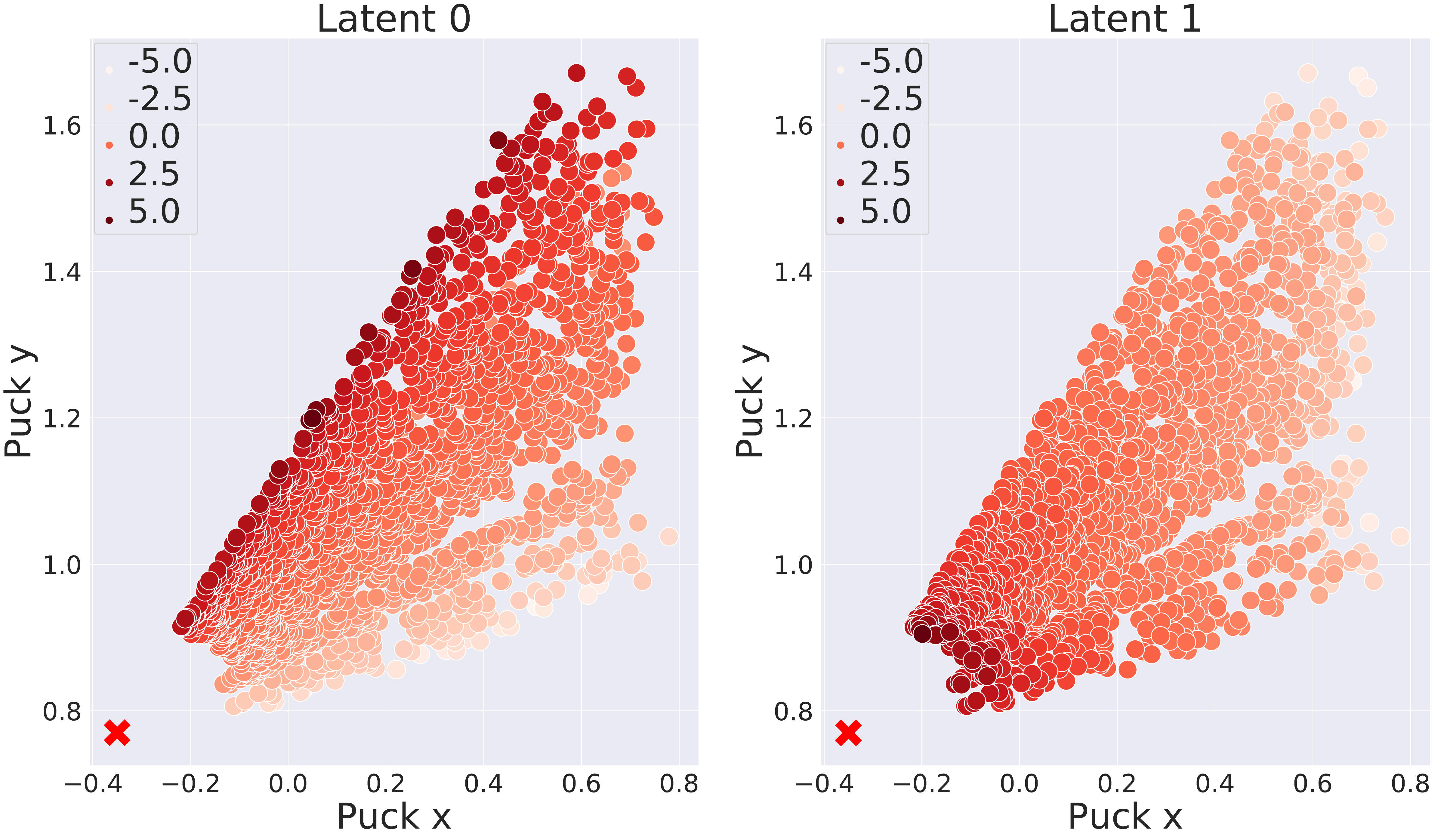}
    \caption{Relation between latent variable and the final position of the hockey puck after executing the corresponding trajectory. Different shades of blue represent values of the latent variables $z_0$ (left) and $z_1$ (right). The red cross represents the initial position of the puck.}
    \label{fig:latent_pos}
\end{figure}

To evaluate the structure of the latent space, we sampled 2000 latent vectors from the latent distribution, executed the corresponding trajectory in the simulator and recorded the final position of the puck.
The results are shown in Figure~\ref{fig:latent_pos}.

The figure illustrates that the model learned to disentangle the hitting angle ($z_0$) from the hitting strength ($z_1$). For example, as the value of $z_0$ increases, the generative model produces trajectories that hit the puck more and more towards the left side in a smooth and continuous manner.

\subsection{Policy training}
The policy is trained in simulation using the simulated setup.
We use $K=16$ rollouts per update and train the policy for $N=3$ adaptation steps.
The policy is represented by a neural network with hidden layer of size 128.
For meta learning, we learn the value of the adaptation step $\alpha$ during training.
The dynamic parameter ranges are shown in Table~\ref{tab:randomizations}.
We use proximal policy optimization (PPO)~\cite{Schulman2017ProximalPO} as the meta optimization algorithm.

As a comparison baseline, we used domain randomization by training a policy by PPO with the same range of randomized dynamics, without adaptation occurring during training.
To provide a fair comparison of the adaptability of initial policy parameters, we used the same adaptation step size $\alpha$ as the trained meta model ($\alpha \approx 0.02$).

We use the following reward function proposed by~\cite{Levine16endtoend}
\begin{equation*}
    r = -d^2 - log(d + b)
\end{equation*}
where $d$ is the distance in meters between the final puck position and the target, and $\alpha$ is a constant (we use $b=10^{-3}$ throughout the experiments).
During policy training, the weights of the trajectory model remained fixed.

\subsection{Simulation experiments}

\begin{figure}
\vspace{6pt}
    \centering
    \begin{subfigure}{0.49\linewidth}
    \centering
    \includegraphics[width=\linewidth]{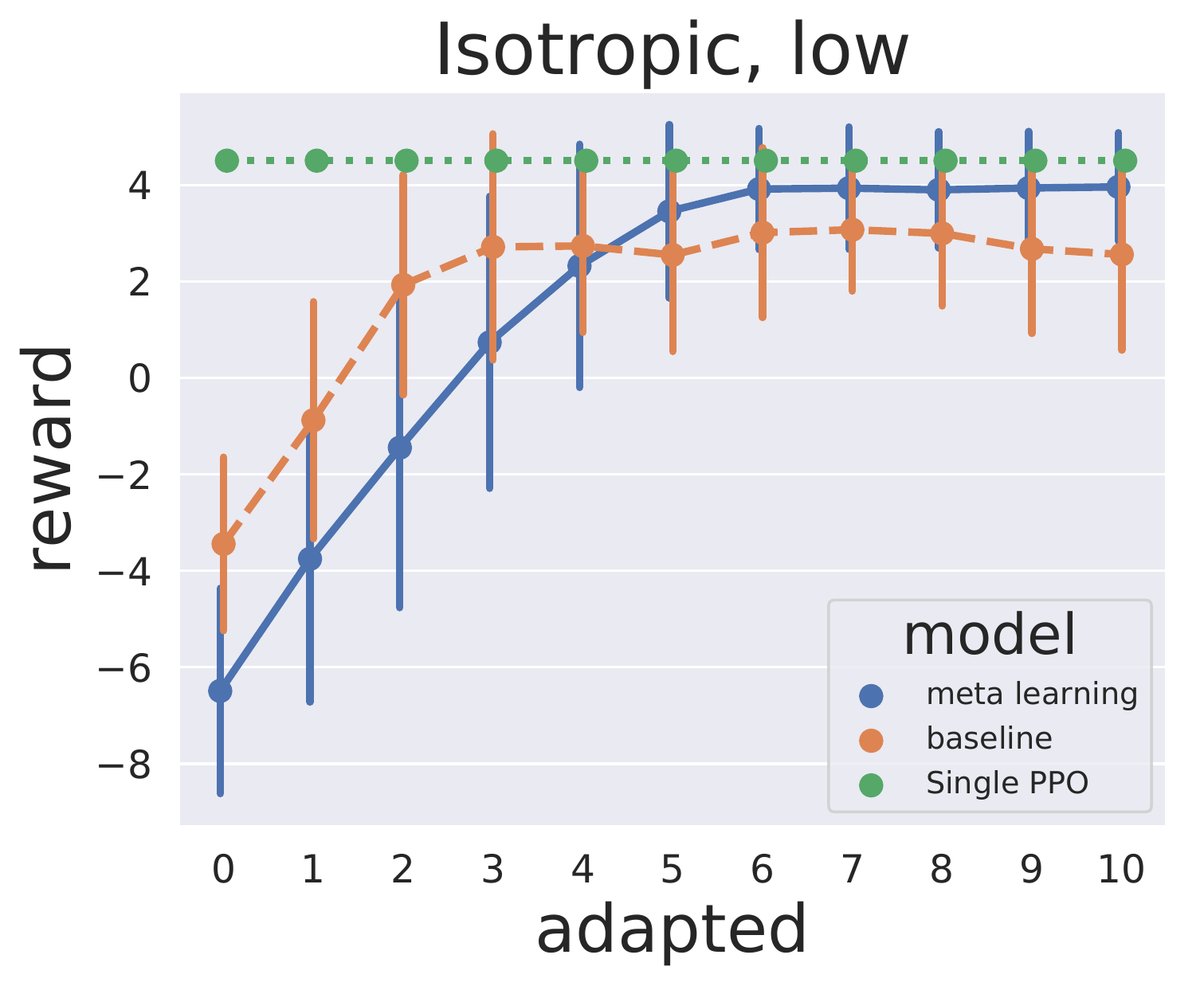}
    \caption{}
    \label{fig:sim_low}
    \end{subfigure}
    \begin{subfigure}{0.49\linewidth}
    \centering
    \includegraphics[width=\linewidth]{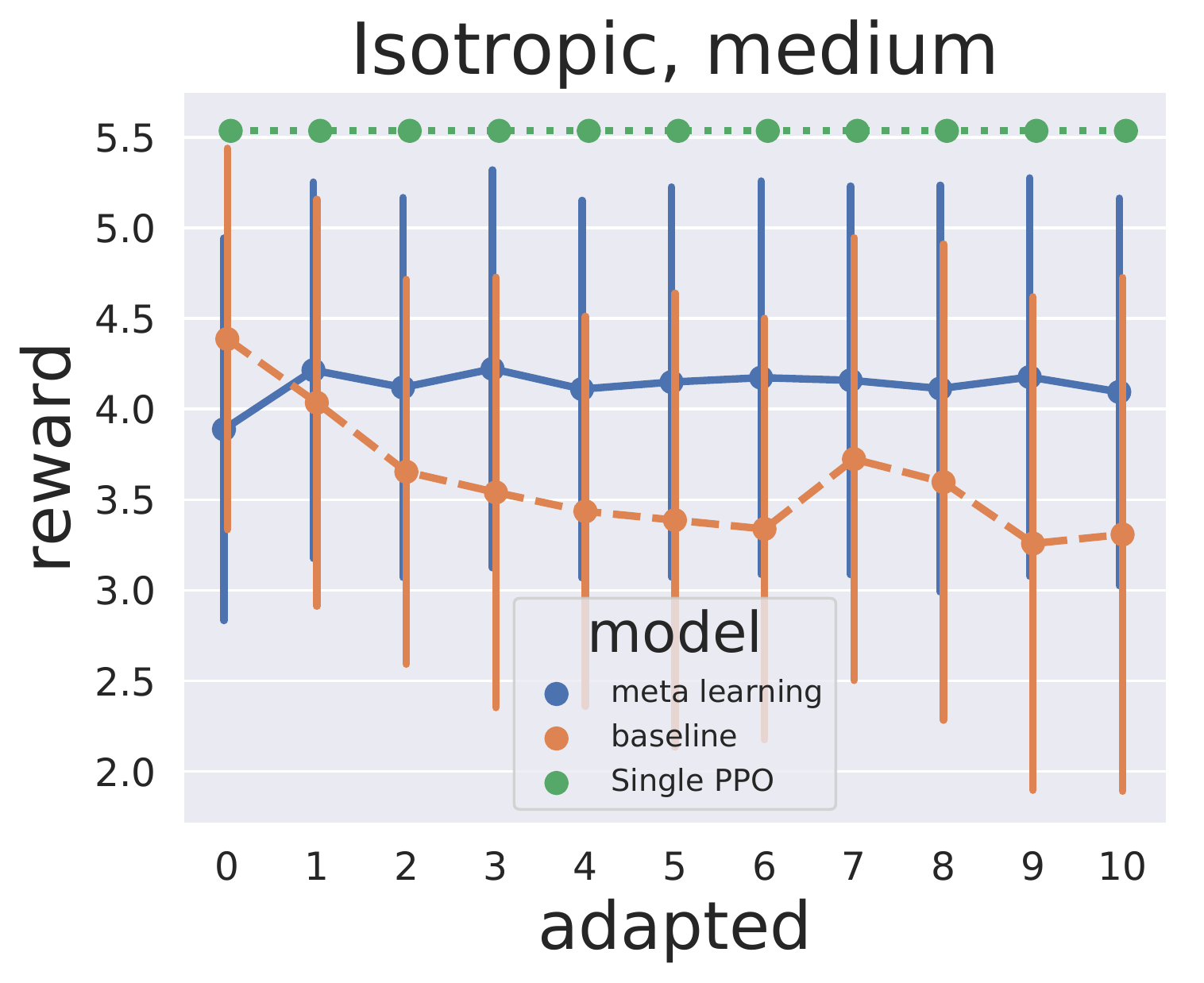}
    \caption{}
    \label{fig:sim_med}
    \end{subfigure}
    \begin{subfigure}{0.49\linewidth}
    \centering
    \includegraphics[width=\linewidth]{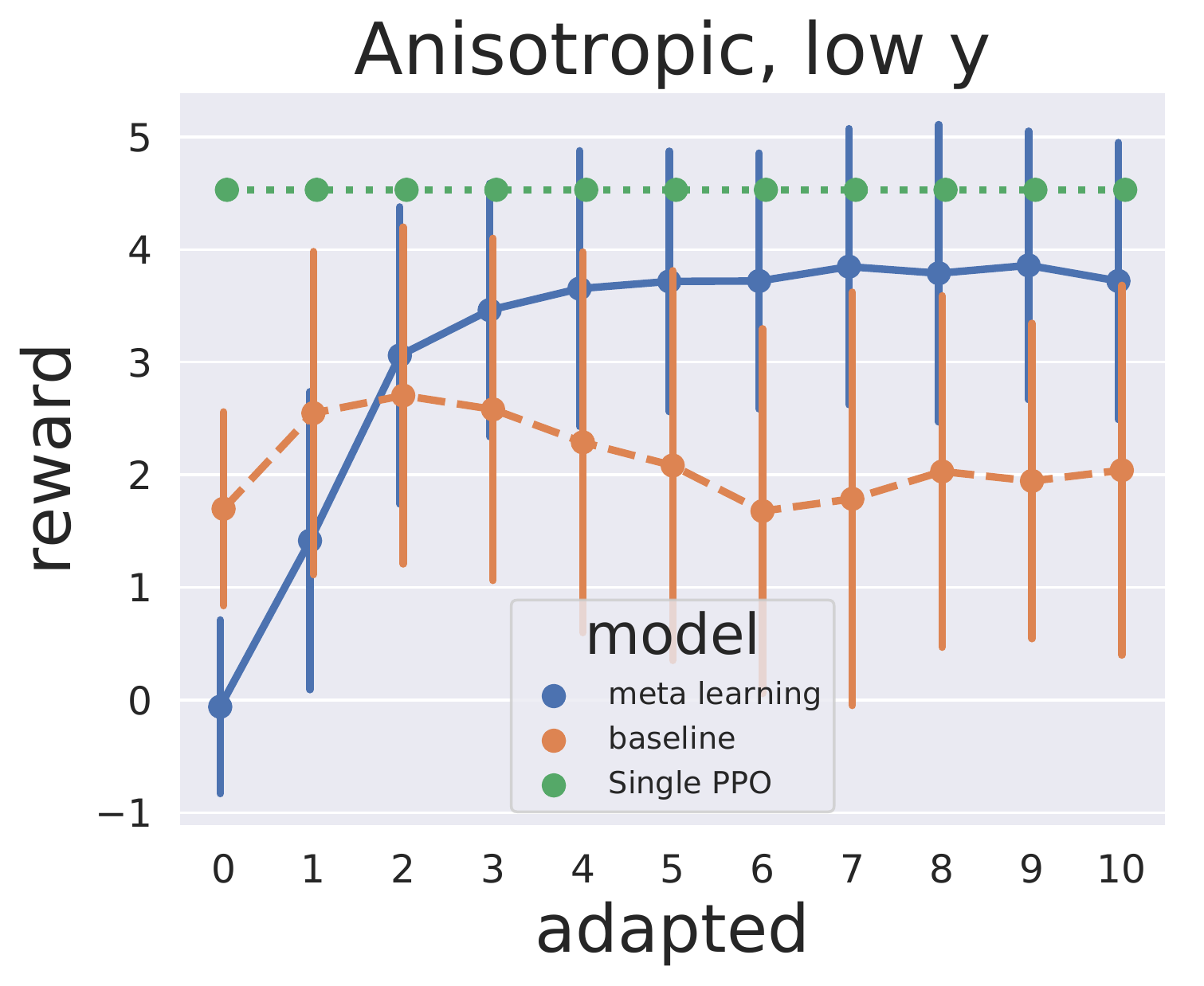}
    \caption{}
    \label{fig:sim_any}
    \end{subfigure}
    \begin{subfigure}{0.49\linewidth}
    \centering
    \includegraphics[width=\linewidth]{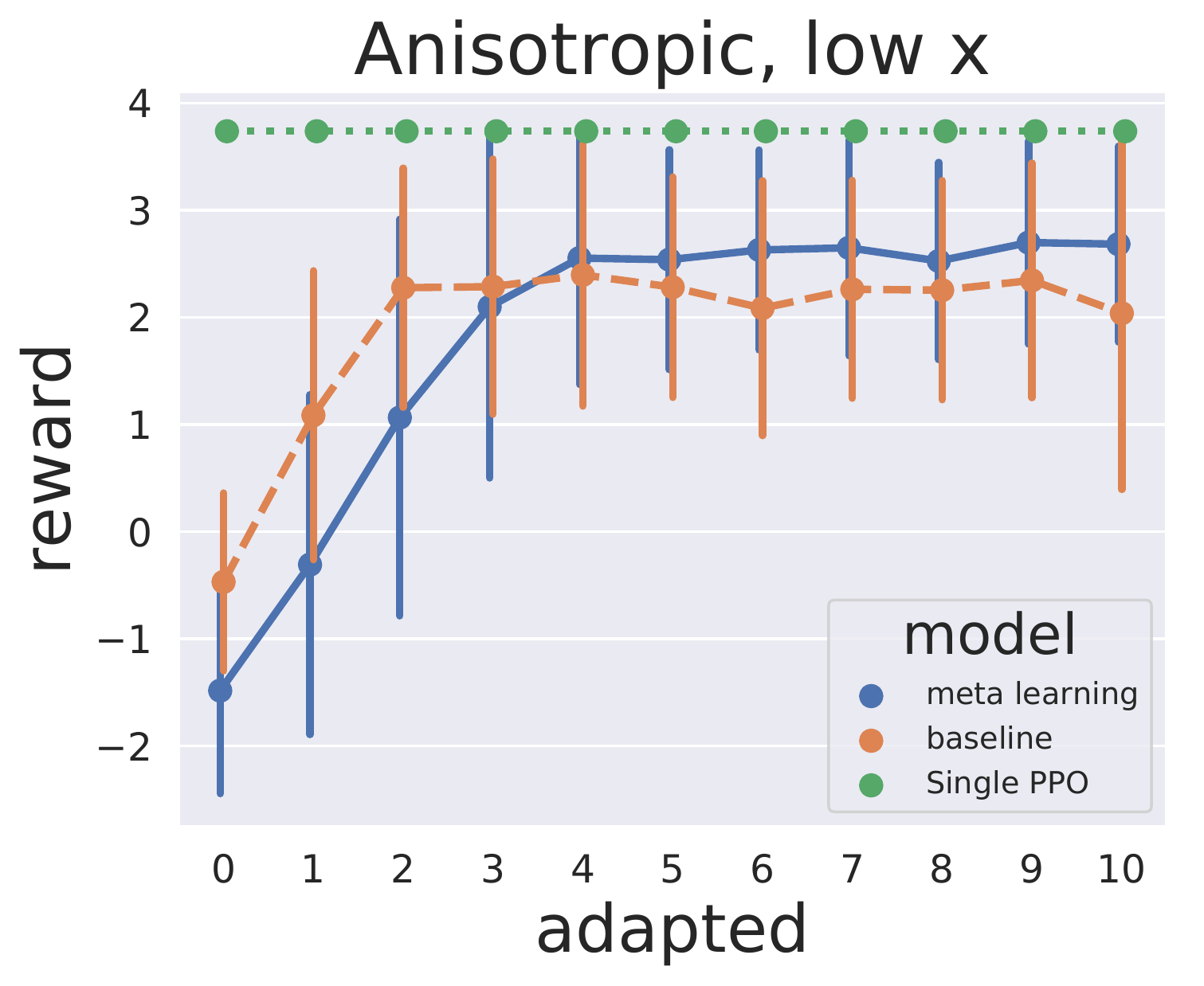}
    \caption{}
    \label{fig:sim_anx}
    \end{subfigure}
    \caption{Comparison of adaptation to different conditions in simulation between our method and the baseline.}
    \label{fig:sim_test}
\end{figure}
\setlength{\tabcolsep}{5pt}
\begin{table}[]
    \centering
    \caption{Dynamic conditions used for simulation experiments}
    \begin{tabular}{c|c|c|c|c|c|c|c|c}
         Experiment & $\mu_x$ & $\mu_y$ & $\mu_\tau$ & $\mu_{rx}$ & $\mu_{ry}$ & $m$ & $\epsilon_x$ & $\epsilon_y$ \\
         \hline
         isotropic, low & 0.15 & 0.15 & 0.01 & 0.1 & 0.1 & 110g & 0.0 & 0.0 \\
         isotr. medium & 0.4 & 0.4 & 0.01 & 0.1 & 0.1 & 110g & 0.0 & 0.0 \\
         anisotr., low x & 0.2 & 0.8 & 0.01 & 0.1 & 0.1 & 110g & 0.0 & 0.0 \\
         anisotr., low y & 0.8 & 0.2 & 0.01 & 0.1 & 0.1 & 110g & 0.0 & 0.0 \\
    \end{tabular}
    \label{tab:sim_fric}
\end{table}
\setlength{\tabcolsep}{6pt}

We studied the proposed method against the baseline in simulation under a variety of dynamic conditions, repeating each experiment 25 times.
To estimate the upper bound on performance for each condition, we also trained a policy for each individual one.
This section illustrates the results of four experiments we consider to be the most unique and interesting: two isotropic friction cases (with low and medium friction) and two anisotropic friction cases with different low frictions directions.
The dynamic parameters used for each of these experiments are shown in Table~\ref{tab:sim_fric}, and the results of this evaluation are presented in Figure~\ref{fig:sim_test}.
Within the selected parameter ranges, changing the mass and the initial position did not have a significant impact on the adaptation performance.

Figure~\ref{fig:sim_test} illustrates that the domain randomization baseline is superior without adaptation, as is expected. However, the proposed method is consistently superior after some adaptation steps. This confirms our initial hypothesis that designing policies specifically for adaptation can increase the adaptation speed and thus help to address domain mismatch. 

Surprisingly, the performance of the domain randomization baseline starts to deteriorate during adaptation in two cases. This most prominent in Fig.~\ref{fig:sim_med}
where the deterioration begins already at the first update. This indicates that the domain randomized policy is somehow unsuitable for adaptation.

\begin{figure}[t]
\vspace{6pt}
    \centering
    \begin{subfigure}{\linewidth}
    \centering
    \includegraphics[width=\linewidth]{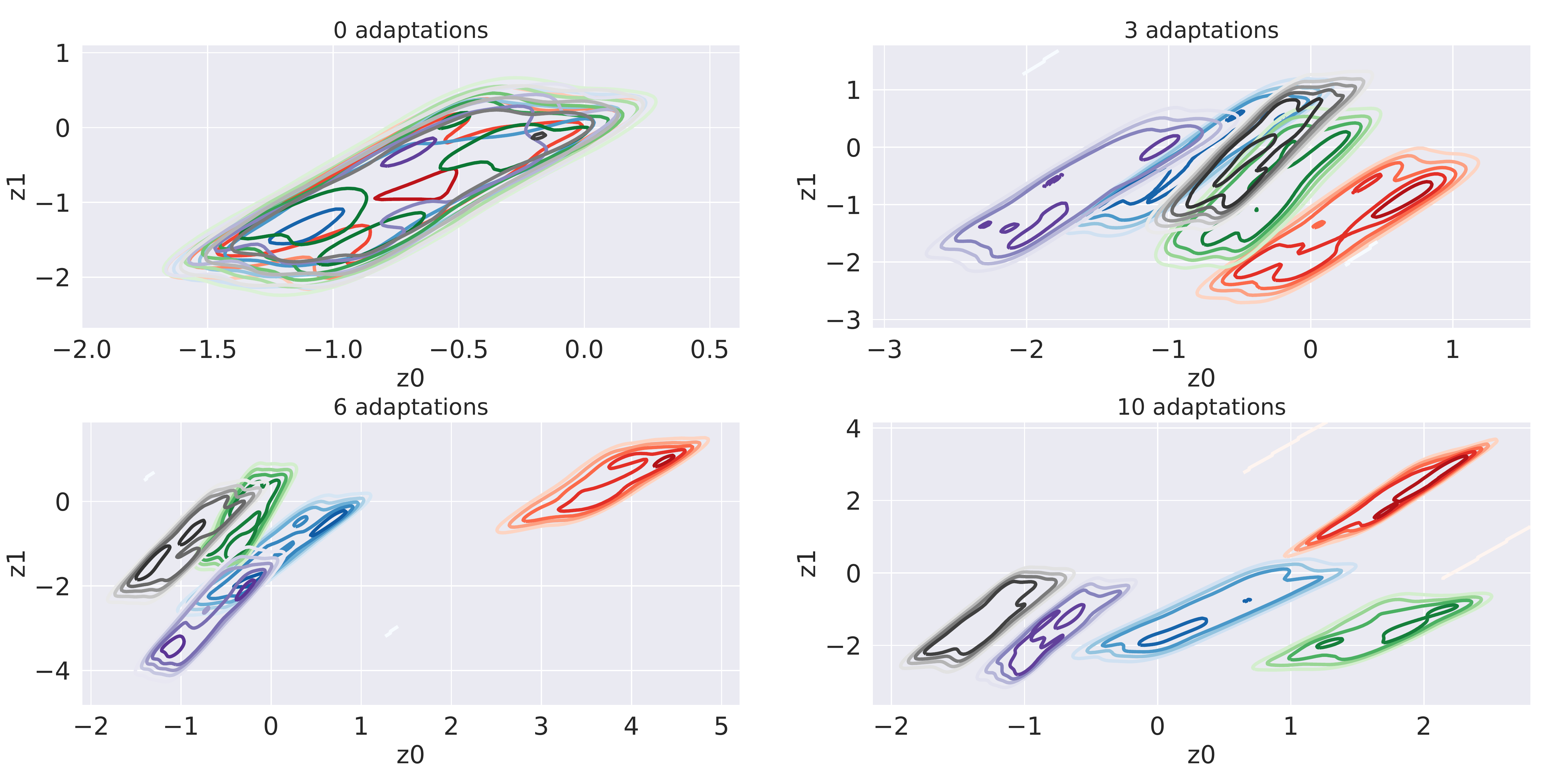}
    \caption{domain randomization baseline}
    \label{fig:latent_base}
    \end{subfigure}
    \begin{subfigure}{\linewidth}
    \centering
    \includegraphics[width=\linewidth]{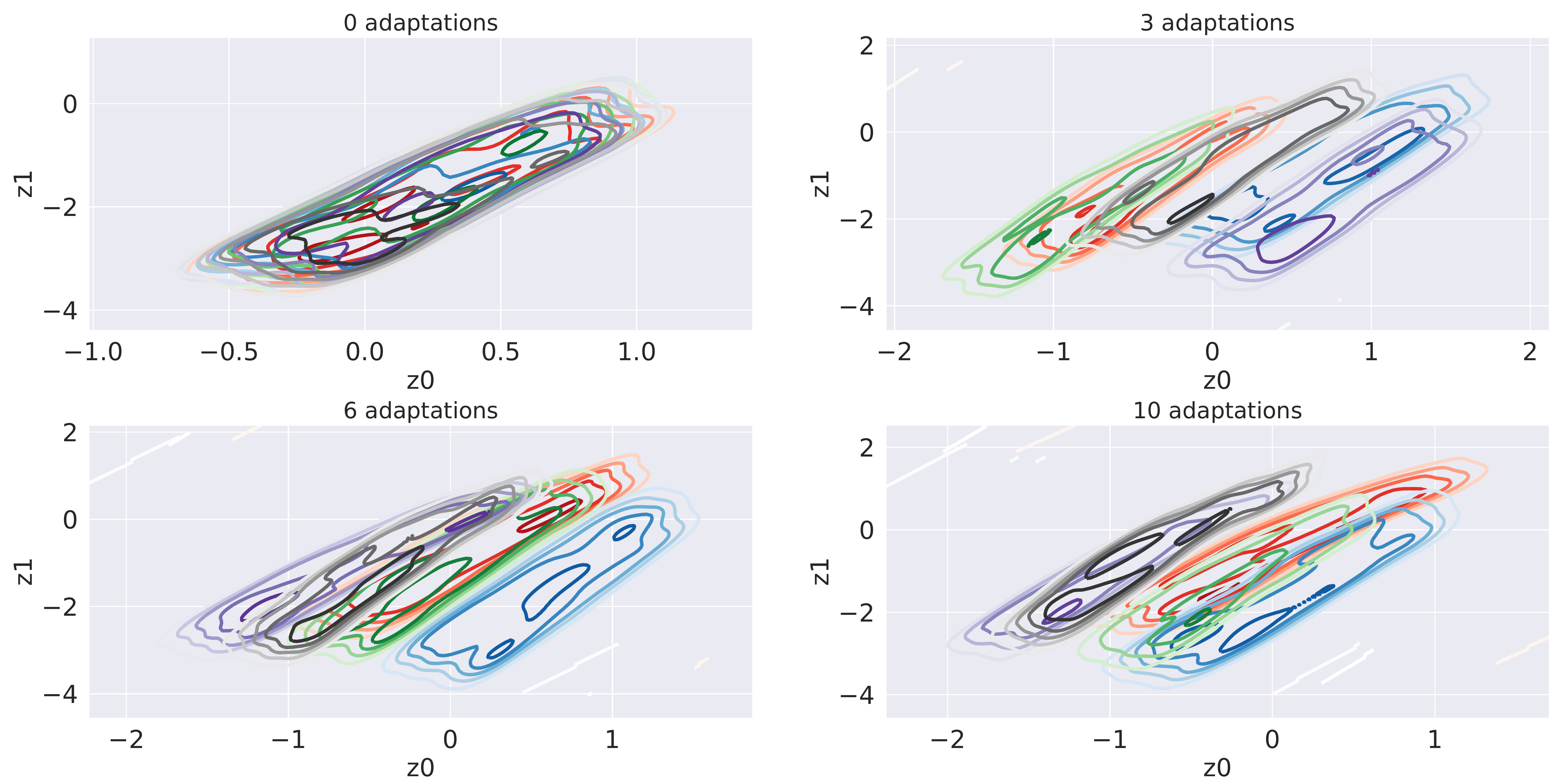}
    \caption{proposed method}
    \label{fig:latent_meta}
    \end{subfigure}
    \caption{Latent action distribution changes over multiple updates, compared across multiple repetitions of the experiment (with each repetition shown in different color). The baseline method provides inconsistent results and is sensitive to inaccuracies in collected samples, while meta learning produces consistent updates.}
    \label{fig:latent_meta_base}
\end{figure}

We analyzed this behaviour further by looking at the latent space action distributions of various repetitions of the experiment.
These distributions are shown in Figure~\ref{fig:latent_meta_base}.
The plots were generated by sampling latent actions for 1000 random goal points at each adaptation step.
Different colours represent different repetitions of the experiment.

Before adaptation, at step 0, the policy takes the same actions during every repetition of the experiment.
However, after a few adaptation steps, the baseline policies start to diverge from each other, as each of them is updated using an independent set of samples.
Performing more adaptations causes these differences to escalate even more. This causes the policy parameters to shift from neighborhood of the origin where the trajectory generator is stable. This in turn is likely to produce more varying trajectories, making the reward gradient used by the updates less stable.

The proposed method does not exhibit such behaviour (Figure~\ref{fig:latent_meta}).
The latent updates follow each other more closely, even after 10 adaptations despite training the meta policy for only 3 adaptation steps.
There are minor differences between the distributions due to random sampling of targets and trajectories but the distributions remain close to the origin and do not diverge.
We hypothesize that is is due to the meta policy being explicitly trained to adapt to new conditions;
it thus learned to perform stable and consistent update steps.
To ensure that this behaviour is a repeatable phenomenon and was not caused by the baseline converging to a bad optimum, we repeated these entire experiments three times, achieving comparable results each time.
Similar findings about regularizing effect of meta learning on policy training were previously reported by Clavera~\cite{clavera2018model}.

After achieving promising results and acquiring deeper understanding of the domain adaptation process with both methods in simulation, we moved on to physical experiments using the previously described experimental setup.

\subsection{Real-world experiments}
We conducted the real world experiments using the setup described in Section~\ref{sec:setup}.
During each run, we conducted 4 adaptations in total.
We evaluated each intermediate policy by taking its mean for 16 randomly chosen target points.
We then sampled another $K=16$ rollouts from the policy to perform an update.
The experiment was repeated 3 times for each hockey puck, resulting in 48 data points for each puck at each adaptation step.

\begin{figure}
\vspace{6pt}
    \centering
    \includegraphics[width=\linewidth]{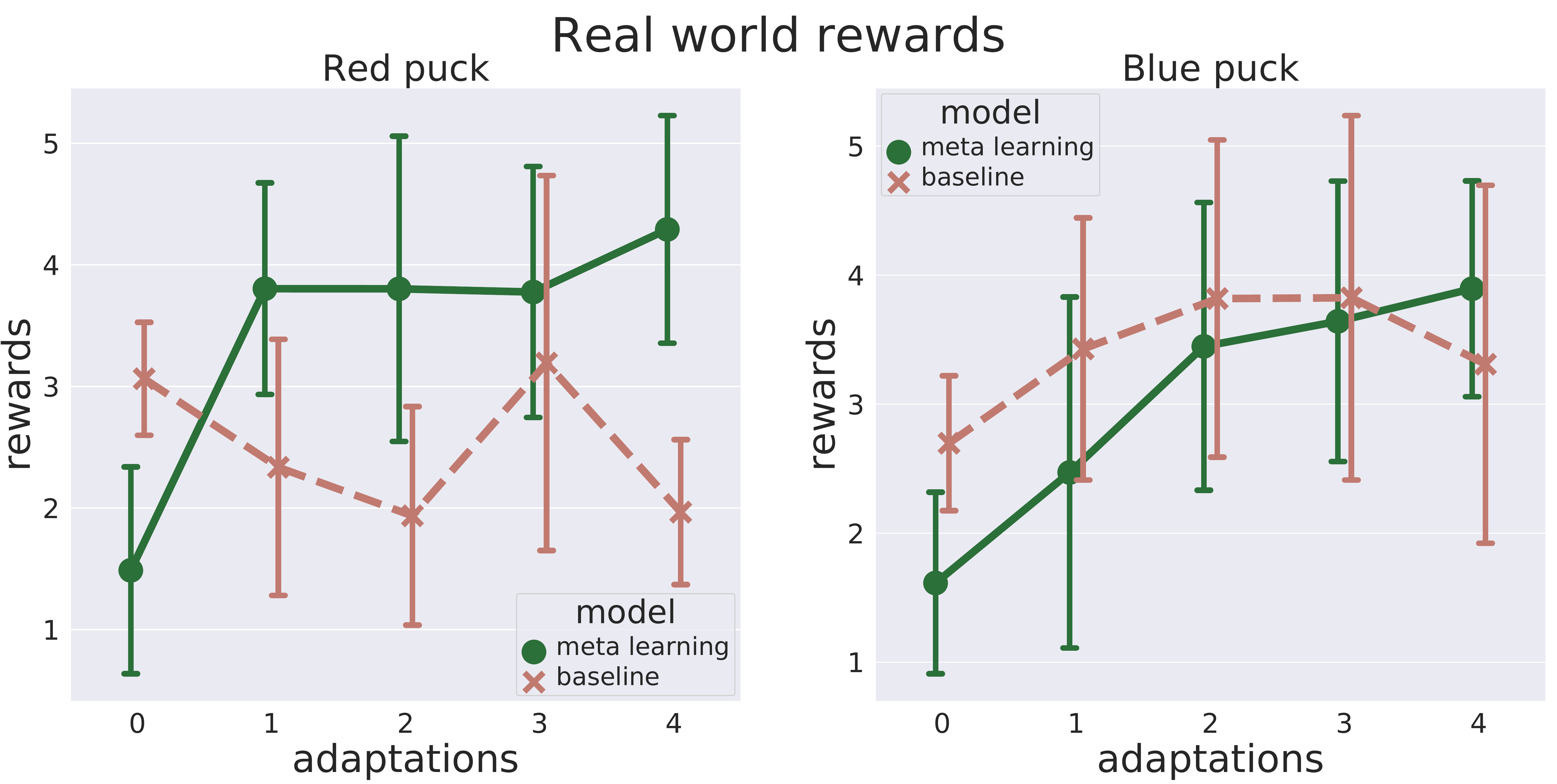}
    \caption{Comparison of real world performance of our method (left) and the baseline (right) with different pucks.}
    \label{fig:real_res}
\end{figure}

The performance comparison between the baseline and our method is shown in Figure~\ref{fig:real_res}.
Consistently with the simulations, the domain randomization baseline (dashed line) is superior without adaptation. Moreover in line with the simulations, the baseline produces inconsistent behaviour during adaptation steps: some policy updates resulted in an overall improvement, while others made the performance deteriorate.
As an extreme case, one of the experiments with the red puck had to be stopped, due to the policy mean landing far enough from the latent distribution such that the trajectory model produced unsuitable trajectories, which was confirmed by studying the latent distribution in a manner similar to the simulation experiments. 
%revealed that this was caused by a policy update resulting in latent vectors far away from the unit normal distribution used to train the trajectory model similar to the red distribution in Figure~\ref{fig:latent_base}.
%In this case, we observed that $z_1$, the latent variable encoding the hitting strength, reached values close to 4, which --- as shown in Figure~\ref{fig:latent_pos} --- corresponds to very gentle hits.
There is a significant difference between the two pucks, with the performance for the blue puck increasing during the first adaptation steps. 
We hypothesize that this is because the low friction of the blue puck provides a stronger policy gradient direction (before adaptation, both policies hit the puck way too strongly, so simply reducing the hitting strength results in a significant increase in rewards).
Nevertheless, additional adaptation steps cause the performance to deteriorate. 
%The baseline, overall, achieves very similar performance with the low-friction blue hockey puck, especially before any adaptations take place.
%This behaviour in low-friction conditions highly resembles the phenomena we've previously observed in simulation.
%We hypothesize that this behaviour is caused by the more `extreme' behaviour of this puck, which results in a clearer policy gradient direction (before adaptation, both policies hit the puck way too strongly, so simply reducing the hitting strength results in a significant increase in rewards).
%However, the baseline fails to keep its performance when more adaptation steps are performed.

The policy trained with meta learning does not suffer from such issues, resulting in consistent adaptation.
The performance either keeps improving or plateaus at a certain level.
This is especially apparent for the higher friction red puck case, where the baseline completely fails to provide any performance improvement whatsoever.
The overall variance is also noticeably smaller than in case of the baseline.
Again, this behaviour is very similar to what was observed in simulation for the medium friction case.

% Conclusions
\section{Conclusions}
\label{sec:conclusions}

In this work, we demonstrated that meta learning can be used as a stable and repeatable simulation-to-real domain adaptation tool.
We observed that adapting a policy trained with standard domain randomization can cause diverse results, with high variance between repetitions and potentially unstable outcomes.
Domain randomization suffered from these issues despite operating in a low dimensional and easy to explore action space.
We also demonstrated that these issues do not exist when the policy is trained for stable adaptation with the proposed meta learning approach.%, showing that a meta learned policy results in a stable and more predictable adaptation.

%Even though the proposed method achieved compelling results and solved some problems of the baseline approach, there are still multiple aspects that could be improved.
When describing the simulated setup, we briefly mentioned how we introduced anisotropic friction to the system to avoid fine tuning contact parameters and make up for unmodeled aspects of the physical setup.
This poses an interesting avenue for further research: whether variations in some parameters can make up for modeling inaccuracies in other physical properties of the system.

In our experiments, we used 16 samples to perform each policy update.
We believe that higher real world sample efficiency could be achieved, especially if the exploration was performed in a more arranged way.
By using gradient based meta learning, we optimize the policy to achieve good performance based on the samples it currently gets, without giving it any incentive to produce useful samples.
While this performed well in our case, further investigation into this could shed some light onto efficient and more informed exploration, potentially leading to higher sample efficiency.

This could potentially be done by off-policy adaptation, that is, gathering samples for policy adaptation from a separate exploration policy different from the adapted one. 
This would allow learning exploratory policies with safety constraints, which could then be used to collect informative samples for quick adaptation.

%===============================================================================

\bibliographystyle{ieeetr}
\bibliography{main}  % .bib

\end{document}